# Online Self-supervised Scene Segmentation for Micro Aerial Vehicles

Shreyansh Daftry, Yashasvi Agrawal and Larry Matthies

*Abstract*— Recently, there have been numerous advances in the development of payload and power constrained lightweight Micro Aerial Vehicles (MAVs). As these robots aspire for high-speed autonomous flights in complex dynamic environments, robust scene understanding at long-range becomes critical. The problem is heavily characterized by either the limitations imposed by sensor capabilities for geometry-based methods, or the need for large-amounts of manually annotated training data required by data-driven methods. This motivates the need to build systems that have the capability to alleviate these problems by exploiting the complimentary strengths of both geometry and data-driven methods. In this paper, we take a step in this direction and propose a generic framework for adaptive scene segmentation using self-supervised online learning. We present this in the context of vision-based autonomous MAV flight, and demonstrate the efficacy of our proposed system through extensive experiments on benchmark datasets and real-world field tests.

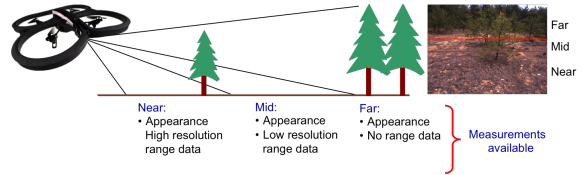

Fig. 1. Near-to-far paradigm. Reliable range information is only available in the near-range, where as appearance information is available even at far-range. We want to exploit the relationship between these disparate modalities, to build a more holistic perception system.

## I. INTRODUCTION

In recent years, autonomous capabilities of resource-constrained autonomous aerial vehicles have seen considerable progress. However, the sensor characteristics of typically used active and passive range sensors on such agile MAVs severely limit the range at which reliable 3D perception can be done. For a typical MAV, this range is usually in the order of $10 \sim 20m$; at longer distances, range data becomes too sparse or noisy for reliable scene understanding from geometry. This presents a challenge, as it makes the system inherently myopic. In contrast, humans effortlessly navigate through most environments - observing, understanding and planning around distant obstacles even in new, previously unseen environments. Such human visual abilities are not solely due to better stereo perception; rather, humans are excellent at reasoning from monocular images. Therefore, in this work we are interested in methods for extending the look-ahead distance of perception systems (See Fig. 1) beyond where geometry can be used, by learning to reason from contextual information in monocular imagery.

The problem of visual recognition has been well studied in general. Recently, learning-based models and features using Convolutional Neural Networks (CNNs) have proven to be more competitive than the traditional geometry-based methods on solving complex vision tasks, including image-based scene segmentation. However, beyond benchmarks and new end-to-end learning applications, they have yet to become the go-to solution for vision-based autonomous navigation. This can be attributed to several reasons: vision algorithms have to contend with continuously evolving, unstructured sensor data during long-term operations. As a result, the performance of data-driven methods do not necessarily translate to real-world scenarios. In addition, most learning methods rely on strongly annotated pixel-accurate data that is highly time-consuming to collect, and often even infeasible. It is the above considerations that motivate our contribution.

In this paper, we advocate that continuous online learning of scene segmentation would allow the system to constantly adapt to its local, and avoids the need to learn a universal scene classifier. This requires a method to automatically generate large amounts of training labels in real-time. Therefore, as proxy, we propose to exploit readily available geometric-cues from the near-range data to generate segmentation labels in real-time, which can then be used to adaptively train the long-range classifier online. Learning from such self-supervision would sidestep the annotation cost to scale up learning performance, and mitigate the challenges of learning algorithms towards long-range perception. We call this self-supervision ability as near-to-far learning.

## II. APPROACH

Near-to-Far learning is a self-supervised learning paradigm that uses visual representations from estimated near-range geometric cues, and thereby learns to reason about far-range scenes. Our system takes as input a RGB image, a depth map, which can be obtained either using an active or passive range sensor, and the estimated pose of the vehicle. We segment the near-range 3D information into obstacle and free-space regions based on the typical application of ground-plane estimation. Assuming the free-space in front of the MAV is locally flat, it is assumed that this plane corresponds locally to the ground. Once the ground plane has been estimated, we use it to segment each pixel for which there is a valid depth information as either obstacle or free-space. This near-range segmentation is then used as a self-supervised training set for the appearance-based learning algorithm.

The problem of learning far-range scene segmentation from appearance can be defined as associating each pixel

S. Daftry and L. Matthies are at Jet Propulsion Laboratory, California Institute of Technology, Pasadena, CA, USA. Y. Agrawal is with the Robotics Institute, Carnegie Mellon University, Pittsburgh, PA, USA. Corresponding Email: Shreyansh.Daftry@jpl.nasa.gov

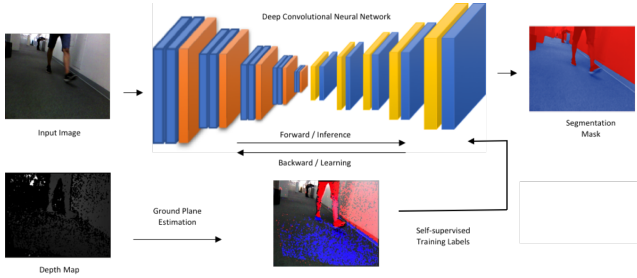

Fig. 2. Schematic overview of the near-to-far scene segmentation approach. Geometric information from the near-range via ground plane estimation is used to generate self-supervised training labels in real-time. These labels are then used to adaptively train a classifier for appearance-based scene segmentation.

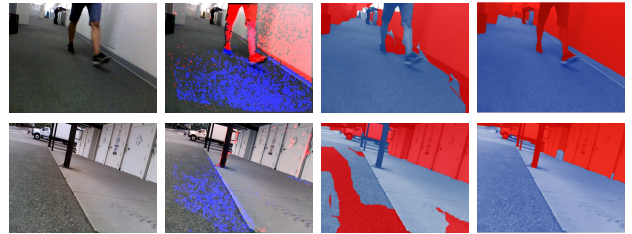

Fig. 3. Qualitative results from MAV flight tests.(Col. 1) Input RGB image, (Col. 2) Near-range geometry-based segmentation overlayed on the input image, (Col. 3) Overlayed segmentation mask for the baseline algorithm (without online learning component) and (Col. 4) Overlayed segmentation mask for the proposed approach. Note: Red is obstacle, blue is free-space.

TABLE I
COMPARISON TO GROUND TRUTH

| Test Dataset | Training Dataset | Method | IoU | AP |
| --- | --- | --- | --- | --- |
| Cityscape | Freiburg | FCN-8s | 71.87 | 77.31 |
| Cityscape | Freiburg | SegNet | 69.43 | 74.67 |
| Cityscape | Freiburg | E-Net | 68.25 | 72.11 |
| Cityscape | Freiburg | Ours (E-Net+online) | **80.36** | **82.21** |

of an input image to one of the semantic classes. Here, we reduce the semantic scene understanding problem to learning a two-class classification: obstacle and free-space. This is the minimum classification required by an autonomous navigation system onboard an MAV to plan collision-free trajectories. Inspired by the recent progress with deep learning, we approach the problem with a CNN that can be trained end-to-end to predict a map of class-labels. While neural network models based-on Fully Convolutional Network [1], have shown exceptional performance on pixel-level segmentation tasks, most of these networks have huge numbers of parameters and long inference times, making them infeasable for robotics applications, which require processing images in real-time on low-latency embedded devices.

In this work, we advocate that reducing the computational burden of semantic segmentation is essential towards making them feasible for deployment on embedded systems for real-world robotics applications. Thus, we design a network based on the recently proposed E-Net architecture [2], that is optimized for both fast inference and high accuracy. E-Net introduces a deep convolutional encoder-decoder model with a bottleneck structure, to build an efficient network architecture that is `18x` faster, has `79x` less parameters but still achieves similar accuracy to prior models. For training the network, we initialized the encoder part of the network with pre-trained weights from a generic dataset (different from the test dataset), in an offline step. The decoder part is initialized using Xavier initialization, and fine-tuned online in real-time by back-propagation using a stochastic gradient descent in a sliding window fashion, where each mini-batch consists of the training set generated from the near-range segmentation over the last $N$ frames.

## III. EXPERIMENTS AND RESULTS

We analyze the qualitative and quantitative performance of our proposed method on benchmark datasets, and demonstrate its efficacy through real-world flight experiments on a MAV. We use an Asctec Hummingbird quad-copter platform, equipped with a NVIDIA Jetson TX1 board, a downward looking camera for state-estimation, and a front-facing Intel RealSense R200 structured light sensor that provides us with visual and range data. During the experiments, the MAV was manually flown by a pilot, while the near-to-far segmentation module was run onboard in real-time as software-in-the-loop.

Fig. 3 shows the qualitative results from some of the flight experiments in a varied set of indoor and outdoor environments, ranging from untextured building corridors to natural cluttered scenes. As baseline, we compare our proposed approach to a system trained only in the supervised manner, without any self-supervised online component. It can be seen that using self-supervised learning in a continuous fashion performs better, especially in scenarios where the environment changes dynamically. For example, in the first example, we notice that as soon as the person appears in near-field, the perception algorithm is able to adapt its learning to classify people as potential obstacles; this is completely missed in the supervised scenario. Furthermore, we perform extensive quantitative evaluations and sensitivity analysis on challenging scene segmentation datasets like Cityscapes and Freiburg Forest, and show the benefits of the proposed online self-supervision (See Table I). Computationally, the geometry-based near-range segmentation module runs at ∼10 hz, and the image-based segmentation module runs at ∼1 hz; this involves both the inference, and the online update step.

In summary, with this work we hope to bridge the gap between geometry-based and data-driven approaches, by taking a step in the direction of building systems that exploit the complimentary benefits of both world.